\documentclass[conference]{IEEEtran}
\usepackage{cite}
\usepackage{siunitx}
\usepackage{multirow}
\usepackage{comment}
\usepackage{amsmath,amssymb,amsfonts}
\usepackage{algorithmic}
\usepackage{graphicx}
\usepackage{textcomp}
\usepackage{xcolor}
\usepackage{subfig}
\def\BibTeX{{\rm B\kern-.05em{\sc i\kern-.025em b}\kern-.08em
    T\kern-.1667em\lower.7ex\hbox{E}\kern-.125emX}}

\begin{document}

\title{Evolutionary Neural Architecture Search for\\Image Restoration}

\author{\IEEEauthorblockN{Gerard Jacques van Wyk}
\IEEEauthorblockA{\textit{Department of Computer Science} \\
\textit{University of Pretoria}\\
Pretoria, South Africa \\
gerard.wyk@gmail.com}
\and
\IEEEauthorblockN{Anna Sergeevna Bosman}
\IEEEauthorblockA{\textit{Department of Computer Science} \\
\textit{University of Pretoria}\\
Pretoria, South Africa \\
annar@cs.up.ac.za}
}

\maketitle

\begin{abstract}
Convolutional neural network (CNN) architectures have traditionally been explored by human experts in a manual search process that is time-consuming and ineffectively explores the massive space of potential solutions. Neural architecture search (NAS) methods automatically search the space of neural network hyperparameters in order to find optimal task-specific architectures. NAS methods have discovered CNN architectures that achieve state-of-the-art performance in image classification among other tasks, however the application of NAS to image-to-image regression problems such as image restoration is sparse. This paper proposes a NAS method that performs computationally efficient evolutionary search of a minimally constrained network architecture search space. The performance of architectures discovered by the proposed method is evaluated on a variety of image restoration tasks applied to the ImageNet64x64 dataset, and compared with human-engineered CNN architectures. The best neural architectures discovered using only 2 GPU-hours of evolutionary search exhibit comparable performance to the human-engineered baseline architecture.
\end{abstract}


\section{Introduction}
It would be hugely inefficient to set many thousands of weight parameters in a modern neural network (NN) by hand, yet many hyperparameters of NN architectures are currently hand-crafted by human experts. Early convolutional neural networks (CNNs) \cite{krizhevsky2012imagenet} contained few hyperparameters, and a small variety of primitive building blocks connected together in simple topologies. In contrast, modern CNN architectures are embedded within an ever-growing search space of possible designs, as researchers continuously invent novel gradient-based optimisation algorithms~\cite{Kingma2014-sm,Hinton2012-xs}, differentiable weight-containing layers~\cite{Hu2017-nn}, activation functions~\cite{Klambauer2017-ur,Clevert2015-rh,He2015-mi}, normalisation methods~\cite{Ioffe2015-ut}, network topology schemes~\cite{Ronneberger2015-xp}, and many other forms of algorithmic and architectural improvements.

As the search space of potential neural architectures grows, it becomes less likely that any given pre-existing network architecture is still the best solution to the problem it was originally designed for. Neural architecture search (NAS) methods attempt to automate the process of finding optimal neural architectures for any given task. NAS methods generally achieve this by treating neural architecture design as an optimisation problem, where the objective is to discover architectures with minimal validation loss for a given task. NAS methods have been demonstrated to be able to discover neural architectures that yield performance comparable or superior to human-engineered neural architectures in the domain of image classification~\cite{Real2018-zf,Pham2018-ad,Real2017-vb,Liu2017-rt,Xie2017-ok} and image restoration~\cite{Suganuma2018-gg}. Most existing NAS methods achieve success by severely limiting the search space of possible solutions.

Image restoration problems are a broad class of image-to-image regression problems where the objective is to reconstruct an original image from a corrupted input. Image restoration problems are of significant scientific and commercial interest. An example scientific application of image restoration is reversal of optical distortions present in optical imaging instruments such as microscopes and telescopes.

This study aims to describe and evaluate a novel NAS method to automate the process of finding optimal CNN architectures for arbitrary image restoration problems. Novel contributions of this work are summarised as follows:
\begin{itemize}
    \item A neural architecture search space is proposed that is both highly expressive and highly searchable.
    \item Adaptive average pooling is effectively employed to eliminate topological constraints.
    \item The feasibility of performing NAS for image-to-image architectures under significant memory and computational time constraints is demonstrated.
\end{itemize}

The rest of the paper is structured as follows: Section~\ref{sec:bg} discusses the background and related work. Section~\ref{sec:searchspace} describes the proposed search space. Section~\ref{sec:evolution} describes the proposed evolutionary algorithm. Section~\ref{sec:exp} discusses the experimental setup. Section~\ref{sec:results} presents the empirical results. Section~\ref{sec:conclusion} summarises the findings of this study, and suggests topics for future research.

\section{Background and Related Work}\label{sec:bg}
Image restoration is a subset of image-to-image regression problems where the objective is to accurately reconstruct an original image, given a corrupted input. Image restoration problems that have been investigated in the context of deep learning include: image denoising~\cite{Nagi2011-rz}, image deblurring~\cite{nah2017deep}, single-image superresolution~\cite{dong2014learning}, and compressive sensing~\cite{mousavi2015deep}, among others. Given a dataset of corrupted and original images, a model can be trained to accept corrupted images as input, and produce restored images as output. The original images are used as target outputs in the process of supervised learning. 

CNNs have a strong prior for the structure of images~\cite{Ulyanov2017-oy}, and have been demonstrated to be highly adept at a wide variety of image processing tasks, including various image restoration tasks~\cite{Nagi2011-rz,nah2017deep,dong2014learning,mousavi2015deep}. Whilst CNNs eliminate the need to design problem-specific image restoration algorithms, they introduce the problem of finding optimal hyperparameter values and designing an appropriate network topology.

To the best of author's knowledge, the only research to date on the topic of NAS methods for image restoration is by Suganuma et al.~\cite{Suganuma2018-gg}, where a neural architecture search space is defined that incorporates a symmetrical convolutional autoencoder architecture constraint in order to reduce the search space complexity. Size mismatches are avoided by allowing skip connections only between layers of the same size. While this search space is highly searchable, it can only represent a very limited section of the true underlying architectural search space. Notable restrictions of their approach include only having ReLU activation functions, only being able to express convolutional autoencoder architectures with skip connections, lack of normalisation layers, and no searchable convolutional layer hyperparameters.

In general, the exisitng NAS approaches~\cite{Real2018-zf,Pham2018-ad,Real2017-vb,Liu2017-rt,Xie2017-ok} have two major limitations:
\begin{enumerate}
    \item A restrictive set of hyperparameter values available to the search algorithm.
    \item A lack of an efficient approach to deal with size and dimensionality mismatches between successive convolutional blocks.
\end{enumerate}

This study proposes an expressive search space by providing an extensive list of hyperparameters and modules available to the evolutionary search algorithm. Additionally, dimensionality constraints are alleviated by applying adaptive average pooling between mismatching modules to enforce valid architectures.

\section{Neural Architecture Search Space}\label{sec:searchspace}  
The neural architecture search space is arguably the core of any NAS method, as it defines the set of all architectures that could possibly be discovered by the search process. Designing a good architecture search space presents a difficult trade-off, where adding more degrees of freedom increases the number of (potentially superior) unique architectures that can be represented, simultaneously increasing the difficulty of the optimisation problem. If the architecture search space is overly constrained, then high performance architectures can not exist within the search space. This section describes the hyperparameters and search space constraints of the proposed neural architecture representation.

\subsection{Network topology representation}
To represent the connected topology of NN architectures, an acyclic directed graph with a single input node and a single output node is used in this study. An adjacency matrix is used to represent the graph. Enforcing a single input and a single output structure is justified in the context of image restoration tasks, as the final model is expected to receive a single image as input and produce a single image as output. 

Each node in the graph constitutes a NN primitive. A primitive is defined as any optionally parameterised differentiable function that receives a tensor input and produces a tensor output. In order to define the network topology as a graph, connective primitives are required that can take multiple previous nodes as input. Previous NAS research generally made use of layer concatenation or elementwise addition. For the proposed architecture representation, the following connective primitives are used: depthwise concatenation, elementwise addition, and elementwise multiplication. Thus, each primitive has one or two predecessor nodes in the adjacency matrix, depending whether it is a single-input primitive or a connective primitive.

The connective operations listed above all share an important constraint: they can only be applied to two inputs of matching dimensionality. This can become a hindrance to an evolutionary algorithm, as crossover and random mutations may introduce dimensionality mismatches, thus generating invalid architectures. This study proposes a novel method to alleviate this constraint: If the shapes of the two input nodes are incompatible, then one of the inputs is resized using adaptive average pooling to ensure compatibility. Adaptive average pooling is simply an average pooling operation that, given an input and output dimensionality, calculates the correct kernel size necessary to produce an output of the given dimensionality from the given input. This simple operation is expected to make the search space significantly more searchable and expressive, since potentially invalid architectures, instead of being discarded, will be converted to valid architectures.

The output of the last executed primitive is taken as the output of the network. This is usually the last primitive in the graph, unless the generated graph exceeded the memory limit, in which case an earlier primitive's output may be used. 

\subsection{Neural network primitives}
A shortcoming of the existing NAS methods is the use of a small variety of unique NN primitives~\cite{Real2018-zf,Pham2018-ad,Real2017-vb,Liu2017-rt,Xie2017-ok,Suganuma2018-gg}. To counter this deficiency, the proposed architecture search space made use of a diverse set of activation functions, normalisation layers, and convolutional layer hyperparameters sourced from recent advancements in neural architecture design. Expanding the set of primitives increases the descriptive power of the network architecture representation at the cost of searchability.

The following activation functions were available to the proposed NAS algorithm: ReLU~\cite{Nair2010-my}, PReLU~\cite{He2015-mi}, ELU~\cite{Clevert2015-rh}, SELU~\cite{Klambauer2017-ur}, hyperbolic tangent (tanh), sigmoid, and softmax.

The following normalisation layers were used: batch normalisation \cite{Ioffe2015-ut}, instance normalisation \cite{Ulyanov2016-ae}, and local response normalisation~\cite{krizhevsky2012imagenet}.

For spatial resolution altering primitives, 2$\times$2 max pooling~\cite{Nagi2011-rz} and nearest neighbour upscaling were employed.

\subsubsection*{Convolutional primitive}
For the convolutional primitive, another divergence was made from previous NAS research. Rather than have several convolutional primitive block types each with pre-set parameters, such as $3\times3$ depthwise convolution block or a $3\times3$ spatial convolution block, this work proposes a single convolutional block type that takes several constrained parameters. The number of input channels is determined by the number of channels in the predecessor node, determined topologically. The number of output channels is constrained to 7 options: same as input channels, 2$\times$input channels, input channels$/2$, 4$\times$input channels, input channels$/4$, 3, and 32. The kernel size of a convolutional layer can be 1$\times$1, 3$\times$3, or 5$\times$5. The stride can be 1 or 2. A convolutional layer can be transposed or regular. A convolutional layer can be a separable depthwise convolution~\cite{chollet2017xception}. Weight normalisation can be set to True or False. The number of parameters of the topological representation is reduced by grouping convolutional blocks together with an activation function and a normalisation layer, since such combination is typically present in human-engineered architectures.

\subsection{Gradient optimisation hyperparameters}
Exisitng NAS methods~\cite{Real2018-zf,Pham2018-ad,Real2017-vb,Liu2017-rt,Xie2017-ok,Suganuma2018-gg} generally do not include gradient optimisation hyperparameters in their NN architecture representations. The proposed approach included a parameter to specify the optimiser type between a choice of Adam~\cite{Kingma2014-sm}, RMSprop~\cite{Hinton2012-xs}, and stochastic gradient descent (SGD) with momentum. A parameter was also included to specify the initial learning rate, constrained to the continuous range of~$[\num{1e-1},\num{1e-5}]$, and a learning rate decay parameter, constrained to the continuous range of $[0,1]$. Optimising the training algorithm for a specific task is important to the success of any NN architecture, thus this minor increase in search space complexity is considered worthwhile.

\subsection{Network constraints}
A number of constraints were imposed on the evolved network representation to limit the search space, and to ensure execution within the predefined computational budget.
\subsubsection*{Memory usage}
During a NN's execution, layers were computed and added to a stack of intermediate tensors that were available as potential inputs to all successive layers as determined by the architecture topology. The execution ended either when the graph was completed, or when memory usage exceeded the set memory limit. In either case, the last tensor in the stack was resized to the target output shape. The memory usage of a NN was estimated as approximately equal to the sum of elements across all the layers at the moment of execution.
\subsubsection*{Time to execute}
If a NN took more than 50 seconds to execute the first 1000 iterations of gradient descent, then gradient optimisation was halted for that individual.

\section{Evolutionary Algorithm for Neural Architecture Search}\label{sec:evolution}
A simple evolutionary algorithm approach was taken. A population of size $N$ was randomly initialised, and the number of allowed gradient iterations was set to $i$. For each generation, all individuals were trained for $i$ iterations on the training set, then the fitness of each individual was evaluated on 1000 minibatches from the validation set. Then, the population was sorted by fitness, and the worst half of the population was killed. If the size of the resulting population was below the minimum population size $n$, the entire population was cloned. Then, $E$ number of elites, or best individuals, were copied directly to the next generation. Crossover operation with probability $p_c = 0.5$ was applied to the rest of the individuals (second parent was randomly chosen from the population), and resulting offspring were used to replace the parents. Uniform crossover was used for the gradient optimisation hyperparameters, and random single-point crossover was used for the primitives. Finally, all individuals were mutated. This process repeated until the predefined computational budget expired.

To promote convergence, an initially large population of $N$ individuals was gradually reduced to the minimal size $n$. Thus, exploration was emphasised at the beginning of the search, with a strong shift towards exploitation at the end of the search.


A high mutation rate of 50\% mutation probability was required to aggressively search the architecture space in the few generations ($<20$) of evolutionary search within the target 2 hour time limit. The following mutation rules were used:

\subsubsection*{Graph}
The adjacency matrix that represents the NN topology was mutated by randomly flipping a bit below the diagonal of the matrix. This mutation randomly connects or disconnects primitives, thus it was possible to generate primitives with no input connections. In this case, the said primitive was connected to the nearest preceding primitive. After the mutations took place, graph pruning was performed to remove primitives with no causal connection to the final output.

\subsubsection*{Primitives}
Network primitives were mutated by adding a primitive, deleting a primitive, or mutating an existing primitive's hyperparameters. Primitive hyperparameters were mutated by being reinitialised on a per-parameter basis according to the mutation probability.

\section{Experimental Setup}\label{sec:exp}
This section describes the experimental setup of the study. Section~\ref{sec:baseline} describes the human-engineered CNN architecture used as the baseline. Section~\ref{sec:data} describes the dataset and the image restoration tasks used in the experiments. Section~\ref{sec:eval} discusses the fitness function used to evaluate the individuals. Section~\ref{sec:enas} lists the hypermarameter values used for the evolutionary search. Section~\ref{sec:comp} lists the hardware and software used to conduct the experiments.

\subsection{Human-engineered baseline network}\label{sec:baseline}
In order to evaluate the performance of the architectures produced by the proposed NAS method, a human-engineered architecture is required as a performance baseline. This baseline NN should ideally be the state-of-the-art architecture for the set of image restoration problems being investigated. For this purpose, a modified version of the U-Net~\cite{Ronneberger2015-xp} architecture was used with PReLU activations, batch normalization, the Adam optimiser, an initial learning rate of $0.01$ halved every 2000 iterations, and a squeeze-and-excitation module~\cite{Hu2017-nn} inserted after batch normalisation in each convolutional block.

For each problem, the baseline CNN was trained for 20,000 iterations using minibatch training with a batch size of 8.

\subsection{Dataset and image restoration tasks}\label{sec:data}
Due to the nature of image restoration problems, any image dataset can be converted to an image restoration dataset by generating input-output pairs required for learning. Corrupted input images are produced by applying a task-specific image degradation function to the original images, and the original images are used as target output. In order to evaluate the performance of the proposed NAS method and the human-engineered architecture, the following image restoration tasks were used as benchmarks: single image superresolution, uniform random noise image denoising, Gaussian random noise image denoising, image deblurring, compressive sensing, and checkerboard rendering reconstruction. 

\subsubsection*{Single image superresolution}
Due to the baseline architecture expecting inputs and outputs to be of the same size, the low-scale input images were resized with nearest neighbour upscaling before being given to the network as input. 
\subsubsection*{Compressive sensing}
Given a random 25\% of image pixels, the network was supposed to reconstruct the missing values.
\subsubsection*{Checkerboard rendering reconstruction} Checkerboard rendering reconstruction is a technique used to optimise real-time graphics upscaling in computer graphics~\cite{checkerboard}. To the best of author's knowledge, this is the first time deep NNs have been used to learn a checkerboard rendering reconstruction filter.

To evaluate the performance of various neural architectures for the above image restoration tasks, an image dataset is required. In choosing a dataset, the following attributes were considered as desirable: a large sample count, a large diversity of natural images, colour images, and a resolution that is large enough to be representative of real world data, while being small enough to quickly train and evaluate a large number of candidate architectures. We eliminated the original version of ImageNet~\cite{Russakovsky2015-lc} for having a resolution that is too large for the given time and memory constraints. Taking all the desirable attributes into account, the ImageNet64x64 dataset~\cite{Chrabaszcz2017-ou} was chosen. The Imagenet64x64 training set consists of 1,281,167 training images divided into 1000 classes. As the name suggests, it is the ImageNet dataset downsampled to a resolution of $64\times64$ pixels.

\subsection{Evaluating the performance of the proposed architectures}\label{sec:eval}
In order to evaluate a network architecture on a given problem, the dataset was split into training, validation, and test sets. The training set was used for gradient-based optimisation of each individual NN, the validation set was used to estimate a NN's performance on unseen images, and a separate test set acted as a measure of performance of a given NN on unseen data. The test set was not observed until the final experiments, or used in any kind of gradient-based or evolutionary optimisation.

For the training/validation/test set split, the ImageNet64x64 dataset was sequentially split into training, validation, and test sets using the ratios of $0.6/0.2/0.2$. The same splits were retained across all experiments.

The chosen loss function across all experiments was the mean squared error (MSE), as it is commonly used for image restoration. In the experimental results, the MSE loss is expressed in the form of peak signal to noise ratio (PSNR). PSNR is a reparameterisation of MSE calculated as: $$\text{PSNR} = 10\log_{10}(1/\text{MSE})$$ 
A higher PSNR indicates higher quality of image restoration. 
\subsection{Evolutionary architecture search}\label{sec:enas}
For each image restoration task, the evolutionary architecture search method trained a population of 32 individuals for 20,000 iterations of gradient-based optimisation per individual on the training set, and evaluated the performance of each individual using 1000 minibatches of size 8 from the validation set. After 2 hours, architecture search was halted, and the performance of the best discovered architecture was evaluated.

\subsection{Hardware and software used}\label{sec:comp}
All experiments were performed on a system with an i7-8700k, a single GTX 1080 ti GPU with 11GB of VRAM, and 16GB of system memory. The dataset was stored on a SSD. The project was implemented in the PyTorch deep learning framework \cite{Paszke2017-hx}.

\section{Results}\label{sec:results} 
This section presents the results of the experiments conducted. The mean PSNR values for all experiments are summarised in Table~\ref{table:results}. 

\begin{table}[b]
\caption{Mean PSNR results for image restoration tasks}
\begin{center}
\begin{tabular}{llll}
\cline{1-4}
\multicolumn{1}{|l|}{Image Restoration Task}                                                                                        & \multicolumn{1}{l|}{Data Subset} & \multicolumn{1}{l|}{\begin{tabular}[c]{@{}l@{}}Baseline \\ PSNR\end{tabular}} & \multicolumn{1}{l|}{\begin{tabular}[c]{@{}l@{}}Evolved \\ PSNR\end{tabular}}  \\ \cline{1-4}
\multicolumn{1}{|l|}{\multirow{3}{*}{\begin{tabular}[c]{@{}l@{}}Single Image Superresolution\\ ($2\times2$ upscaling)\end{tabular}}}       & \multicolumn{1}{l|}{Training}      & \multicolumn{1}{l|}{23.7278}                                                             & \multicolumn{1}{l|}{23.7280}                                                            \\ \cline{2-4}
\multicolumn{1}{|l|}{}                                                                                                              & \multicolumn{1}{l|}{Validation}    & \multicolumn{1}{l|}{23.7251}                                                             & \multicolumn{1}{l|}{23.7336}                                                             \\ \cline{2-4}
\multicolumn{1}{|l|}{}                                                                                                              & \multicolumn{1}{l|}{Test}          & \multicolumn{1}{l|}{23.7315}                                                             & \multicolumn{1}{l|}{23.7321}                                                            \\ \cline{1-4}
\multicolumn{1}{|l|}{\multirow{3}{*}{\begin{tabular}[c]{@{}l@{}}Denoising\\ (Uniform noise $\sim[-0.5, 0.5]$)\end{tabular}}}             & \multicolumn{1}{l|}{Training}      & \multicolumn{1}{l|}{23.5569}                                                             & \multicolumn{1}{l|}{21.3105}                                                            \\ \cline{2-4}
\multicolumn{1}{|l|}{}                                                                                                              & \multicolumn{1}{l|}{Validation}    & \multicolumn{1}{l|}{23.5585}                                                             & \multicolumn{1}{l|}{21.3039}                                                             \\ \cline{2-4}
\multicolumn{1}{|l|}{}                                                                                                              & \multicolumn{1}{l|}{Test}          & \multicolumn{1}{l|}{23.5605}                                                             & \multicolumn{1}{l|}{21.3073}                                                            \\ \cline{1-4}
\multicolumn{1}{|l|}{\multirow{3}{*}{\begin{tabular}[c]{@{}l@{}}Denoising\\ (Gaussian noise, $\sigma=0.2$)\end{tabular}}}                & \multicolumn{1}{l|}{Training}      & \multicolumn{1}{l|}{23.6653}                                                             & \multicolumn{1}{l|}{21.7955}                                                            \\ \cline{2-4}
\multicolumn{1}{|l|}{}                                                                                                              & \multicolumn{1}{l|}{Validation}    & \multicolumn{1}{l|}{23.6611}                                                             & \multicolumn{1}{l|}{21.7894}                                                             \\ \cline{2-4}
\multicolumn{1}{|l|}{}                                                                                                              & \multicolumn{1}{l|}{Test}          & \multicolumn{1}{l|}{23.6656}                                                             & \multicolumn{1}{l|}{21.7929}                                                            \\ \cline{1-4}

\multicolumn{1}{|l|}{\multirow{3}{*}{\begin{tabular}[c]{@{}l@{}}Deblurring\\ (Gaussian kernel, $\sigma=2, r=3$)\end{tabular}}} & \multicolumn{1}{l|}{Training}      & \multicolumn{1}{l|}{25.7073}                                                             & \multicolumn{1}{l|}{23.8966}                                                            \\ \cline{2-4}
\multicolumn{1}{|l|}{}                                                                                                              & \multicolumn{1}{l|}{Validation}    & \multicolumn{1}{l|}{25.7085}                                                             & \multicolumn{1}{l|}{23.8922}                                                             \\ \cline{2-4}
\multicolumn{1}{|l|}{}                                                                                                              & \multicolumn{1}{l|}{Test}          & \multicolumn{1}{l|}{25.7142}                                                             & \multicolumn{1}{l|}{23.8916}                                                            \\ \cline{1-4}
\multicolumn{1}{|l|}{\multirow{3}{*}{Compressive Sensing}}                                                                          & \multicolumn{1}{l|}{Training}      & \multicolumn{1}{l|}{27.0249}                                                             & \multicolumn{1}{l|}{21.7724}                                                            \\ \cline{2-4}
\multicolumn{1}{|l|}{}                                                                                                              & \multicolumn{1}{l|}{Validation}    & \multicolumn{1}{l|}{27.0210}                                                             & \multicolumn{1}{l|}{21.7687}                                                             \\ \cline{2-4}
\multicolumn{1}{|l|}{}                                                                                                              & \multicolumn{1}{l|}{Test}          & \multicolumn{1}{l|}{27.0217}                                                             & \multicolumn{1}{l|}{21.7712}                                                            \\ \cline{1-4}
\multicolumn{1}{|l|}{\multirow{3}{*}{\begin{tabular}[c]{@{}l@{}}Checkerboard Rendering\\ Reconstruction\end{tabular}}}              & \multicolumn{1}{l|}{Training}      & \multicolumn{1}{l|}{26.9629}                                                             & \multicolumn{1}{l|}{25.6542}                                                            \\ \cline{2-4}
\multicolumn{1}{|l|}{}                                                                                                              & \multicolumn{1}{l|}{Validation}    & \multicolumn{1}{l|}{26.9642}                                                             & \multicolumn{1}{l|}{25.6549}                                                             \\ \cline{2-4}
\multicolumn{1}{|l|}{}                                                                                                              & \multicolumn{1}{l|}{Test}          & \multicolumn{1}{l|}{26.9637}                                                             & \multicolumn{1}{l|}{25.6539}                                                            \\ \hline
\end{tabular}
\label{table:results}
\end{center}
\end{table}

\begin{figure*}
	\centering
	\subfloat[{Superresolution: Baseline}] {
	    \includegraphics[width=0.45\linewidth]{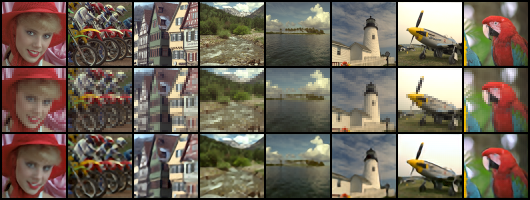}
	}
	\subfloat[{Superresolution: Evolved}] {
	    \includegraphics[width=0.45\linewidth]{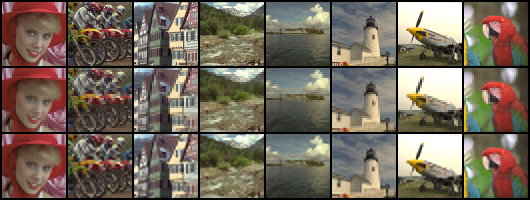}
	}\\
	\subfloat[{Image denoising - Uniform random noise: Baseline}] {
	    \includegraphics[width=0.45\linewidth]{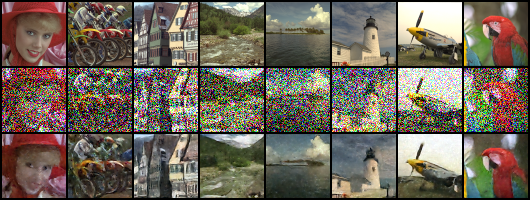}
	}
	\subfloat[{Image denoising - Uniform random noise: Evolved}] {
	    \includegraphics[width=0.45\linewidth]{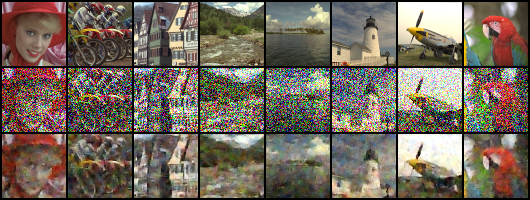}
	}\\
	\subfloat[{Image denoising - Gaussian random noise: Baseline}] {
	    \includegraphics[width=0.45\linewidth]{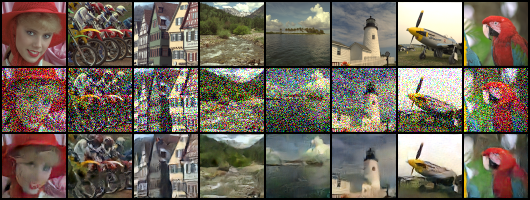}
	}
	\subfloat[{Image denoising - Gaussian random noise: Evolved}] {
	    \includegraphics[width=0.45\linewidth]{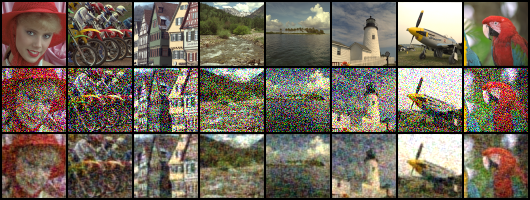}
	}
	
  \caption{Visual comparison of the baseline and evolved NN architecture performance. Top rows: ground truth, middle rows: corrupted input, bottom rows: neural network output.}\label{fig:comparison1}
\end{figure*}

\begin{figure*}
	\centering
	\subfloat[{Image deblurring: Baseline}] {
	    \includegraphics[width=0.45\linewidth]{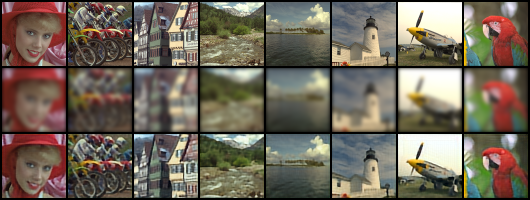}
	}
	\subfloat[{Image deblurring: Evolved}] {
	    \includegraphics[width=0.45\linewidth]{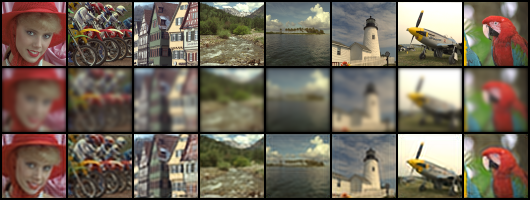}
	}\\
	\subfloat[{Compressive sensing: Baseline}] {
	    \includegraphics[width=0.45\linewidth]{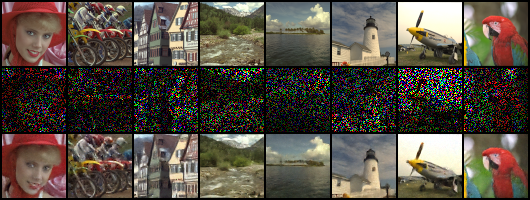}
	}
	\subfloat[{Compressive sensing: Evolved}] {
	    \includegraphics[width=0.45\linewidth]{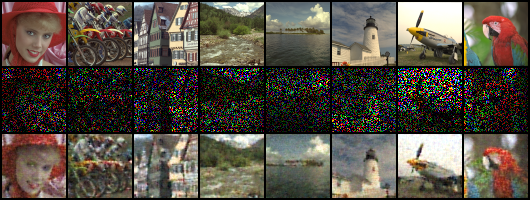}
	}\\
	\subfloat[{Checkerboard rendering reconstruction: Baseline}] {
	    \includegraphics[width=0.45\linewidth]{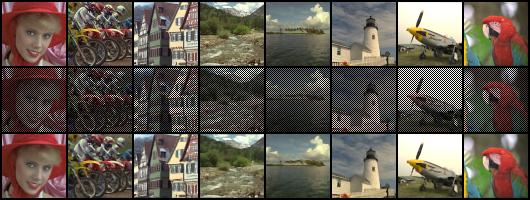}
	}
	\subfloat[{Checkerboard rendering reconstruction: Evolved}] {
	    \includegraphics[width=0.45\linewidth]{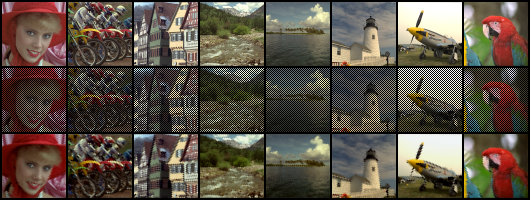}
	}
  \caption{Visual comparison of the baseline and evolved NN architecture performance. Top rows: ground truth, middle rows: corrupted input, bottom rows: neural network output.}\label{fig:comparison2}
\end{figure*}

It is evident from the PSNR values that the human-engineered architecture performed better than the best evolved architecture on most image restoration problems, with the exception of the single image superresolution, where both architectures performed on par. However, the evolved architectures performed on a comparable level, which is impressive given the time and complexity constraints imposed on the evolutionary process. While the human-engineered architecture was heavily overparameterised, the evolved architectures were forced to learn to perform the same task with a significantly smaller number of total parameters.  Fig.~\ref{fig:comparison1} and~\ref{fig:comparison2} illustrate the performance of the human-engineered and the evolved architectures on the six image restoration tasks applied to a set of 8 images selected from the publicly available Kodak  image suite~\cite{franzen1999kodak}. It is evident from the visual inspection that the evolved architectures performed adequately. In multiple cases, the difference in performance is barely noticeable.

Another interesting result is the tight coupling between the training, validation, and test set accuracies, i.e. very little to no overfitting. The minimal variance between the dataset splits is likely due to the large sample count of the dataset, and the good generalisation ability of the network architectures investigated.

\subsection{Properties of evolved architectures}
Example evolved architecture parameters are presented in Tables~\ref{table:super}, \ref{table:gauss}, \ref{table:compressive}, and~\ref{table:checkerboard}. PyTorch syntax is used to describe the primitives.

Table~\ref{table:super} shows one of the larger evolved architectures, with a total of 14 nodes. Multiple convolutional blocks of varying dimensionality were evolved, and combined using adaptive average pooling. This architecture performed as well as the human engineered architecture, but was smaller in size.

Table~\ref{table:gauss} shows that for the denoising task, a very compact CNN architecture emerged. Same applies to the checkerboard reconstruction, shown in Table~\ref{table:checkerboard}. Both have employed concatenation of the earlier layer signals with the later layer signals, similar to the U-Net architecture.

Table~\ref{table:compressive} shows an example of a simple feed-forward architecture evolved for the compressive sensing task. In this case, no concatenations took place, and sigmoidal functions were used in a number of layers. Compressive sensing task provided only 25\% of valid inputs, which may have caused concatenation of the input signals to the hidden layer signals to be ineffective.

It is hard to draw definite conclusions from the small sample size of high performance architectures discovered by the proposed NAS method, but certain tendencies can be observed nevertheless. The high frequency of Adam and RMSprop as opposed to SGD training, the high frequency of transposed convolutional layers, and the high frequency of rectifier-based activation functions are in line with the existing best practices among CNN practitioners. Perhaps the most interesting property of the evolved architectures is the sheer diversity of network configurations that were produced. This serves as evidence for the expressivity of the proposed neural architecture search space.

\begin{table*}[tb]
\caption{Parameters of the best evolved architecture for the single image superresolution task.}
\label{table:super}
\resizebox{\linewidth}{!}{%
\begin{tabular}{lll}
\multicolumn{3}{l}{Optimizer: Adam }\\
\multicolumn{3}{l}{Initial learning rate: 0.041888 }\\
\multicolumn{3}{l}{Learning rate decay factor: 0.136235 } \\ \hline
\multicolumn{1}{|l|}{Node index} & \multicolumn{1}{l|}{Input Node(s)} & \multicolumn{1}{l|}{Node type} \\ \hline
\multicolumn{1}{|l|}{0} & \multicolumn{1}{l|}{} & \multicolumn{1}{l|}{Input node} \\ \hline
\multicolumn{1}{|l|}{1} & \multicolumn{1}{l|}{0} & \multicolumn{1}{l|}{\begin{tabular}[c]{@{}l@{}}(conv): ConvTranspose2d(3, 12, kernel\_size=(3, 3), stride=(2, 2), padding=(1, 1), output\_padding=(1, 1), groups=3, bias=False)\\ (activ): PReLU(num\_parameters=1)\\ (norm): BatchNorm2d(12, eps=1e-05, momentum=0.1, affine=True, track\_running\_stats=True)\end{tabular}} \\ \hline
\multicolumn{1}{|l|}{2} & \multicolumn{1}{l|}{1} & \multicolumn{1}{l|}{upsample} \\ \hline
\multicolumn{1}{|l|}{3} & \multicolumn{1}{l|}{0, 2} & \multicolumn{1}{l|}{mul - resize to first} \\ \hline
\multicolumn{1}{|l|}{4} & \multicolumn{1}{l|}{3} & \multicolumn{1}{l|}{\begin{tabular}[c]{@{}l@{}}(conv): Conv2d(3, 32, kernel\_size=(3, 3), stride=(1, 1), padding=(1, 1))\\ (activ): SELU()\end{tabular}} \\ \hline
\multicolumn{1}{|l|}{5} & \multicolumn{1}{l|}{4} & \multicolumn{1}{l|}{\begin{tabular}[c]{@{}l@{}}(conv): ConvTranspose2d(32, 16, kernel\_size=(3, 3), stride=(2, 2), padding=(1, 1), output\_padding=(1, 1))\\ (activ): SELU()\\ (norm): LocalResponseNorm(16, alpha=0.0001, beta=0.75, k=1)\end{tabular}} \\ \hline
\multicolumn{1}{|l|}{6} & \multicolumn{1}{l|}{0, 5} & \multicolumn{1}{l|}{add - resize to second} \\ \hline
\multicolumn{1}{|l|}{7} & \multicolumn{1}{l|}{1} & \multicolumn{1}{l|}{\begin{tabular}[c]{@{}l@{}}(conv): ConvTranspose2d(12, 48, kernel\_size=(3, 3), stride=(2, 2), padding=(1, 1), output\_padding=(1, 1), bias=False)\\ (activ): ReLU()\\ (norm): InstanceNorm2d(48, eps=1e-05, momentum=0.1, affine=False, track\_running\_stats=False)\end{tabular}} \\ \hline
\multicolumn{1}{|l|}{8} & \multicolumn{1}{l|}{7} & \multicolumn{1}{l|}{upsample} \\ \hline
\multicolumn{1}{|l|}{9} & \multicolumn{1}{l|}{7, 8} & \multicolumn{1}{l|}{mul - resize to second} \\ \hline
\multicolumn{1}{|l|}{10} & \multicolumn{1}{l|}{7, 9} & \multicolumn{1}{l|}{add - resize to second} \\ \hline
\multicolumn{1}{|l|}{11} & \multicolumn{1}{l|}{10} & \multicolumn{1}{l|}{\begin{tabular}[c]{@{}l@{}}(conv): Conv2d(48, 3, kernel\_size=(3, 3), stride=(2, 2), padding=(1, 1), bias=False)\\ (activ): SELU()\\ (norm): Softmax2d()\end{tabular}} \\ \hline
\multicolumn{1}{|l|}{12} & \multicolumn{1}{l|}{6, 11} & \multicolumn{1}{l|}{mul - resize to first} \\ \hline
\multicolumn{1}{|l|}{13} & \multicolumn{1}{l|}{12} & \multicolumn{1}{l|}{\begin{tabular}[c]{@{}l@{}}(conv): ConvTranspose2d(16, 8, kernel\_size=(3, 3), stride=(2, 2), padding=(1, 1), output\_padding=(1, 1))\\ (activ): SELU()\\ (norm): BatchNorm2d(8, eps=1e-05, momentum=0.1, affine=True, track\_running\_stats=True)\end{tabular}} \\ \hline
\multicolumn{1}{|l|}{14} & \multicolumn{1}{l|}{13, 13} & \multicolumn{1}{l|}{add - resize to first} \\ \hline
\end{tabular}%
}
\end{table*}

\begin{table*}[tbh]
\caption{Parameters of the best evolved architecture for the image denoising (Gaussian noise) task.}
\label{table:gauss}
\resizebox{\linewidth}{!}{%
\begin{tabular}{lll}
\multicolumn{3}{l}{Optimizer: RMSprop }\\
\multicolumn{3}{l}{Initial learning rate: 0.038556 }\\
\multicolumn{3}{l}{Learning rate decay factor: 0.133418 }\\ \hline
\multicolumn{1}{|l|}{Node index} & \multicolumn{1}{l|}{Input Node(s)} & \multicolumn{1}{l|}{Node type} \\ \hline
\multicolumn{1}{|l|}{0} & \multicolumn{1}{l|}{} & \multicolumn{1}{l|}{Input node} \\ \hline
\multicolumn{1}{|l|}{1} & \multicolumn{1}{l|}{0} & \multicolumn{1}{l|}{\begin{tabular}[c]{@{}l@{}}(conv): Conv2d(3, 12, kernel\_size=(5, 5), stride=(2, 2), padding=(2, 2), groups=3, bias=False)\\ (activ): PReLU(num\_parameters=1)\end{tabular}} \\ \hline
\multicolumn{1}{|l|}{2} & \multicolumn{1}{l|}{1} & \multicolumn{1}{l|}{\begin{tabular}[c]{@{}l@{}}(conv): ConvTranspose2d(12, 3, kernel\_size=(3, 3), stride=(2, 2), padding=(1, 1), output\_padding=(1, 1), bias=False)\\ (activ): Tanh()\\ (norm): LocalResponseNorm(3, alpha=0.0001, beta=0.75, k=1)\end{tabular}} \\ \hline
\multicolumn{1}{|l|}{3} & \multicolumn{1}{l|}{0} & \multicolumn{1}{l|}{(conv): Conv2d(3, 1, kernel\_size=(3, 3), stride=(1, 1), padding=(1, 1))} \\ \hline
\multicolumn{1}{|l|}{4} & \multicolumn{1}{l|}{1, 3} & \multicolumn{1}{l|}{concat - resize to second} \\ \hline
\multicolumn{1}{|l|}{5} & \multicolumn{1}{l|}{2, 3} & \multicolumn{1}{l|}{concat - resize to first} \\ \hline
\multicolumn{1}{|l|}{6} & \multicolumn{1}{l|}{4} & \multicolumn{1}{l|}{\begin{tabular}[c]{@{}l@{}}(conv): ConvTranspose2d(13, 3, kernel\_size=(3, 3), stride=(2, 2), padding=(1, 1), output\_padding=(1, 1))\\ (activ): ELU(alpha=1.0)\\ (norm): LocalResponseNorm(3, alpha=0.0001, beta=0.75, k=1)\end{tabular}} \\ \hline
\end{tabular}%
}
\end{table*}

\begin{table*}[tbh]
\caption{Parameters of the best evolved architecture for the compressive sensing task.}
\label{table:compressive}
\resizebox{\linewidth}{!}{%
\begin{tabular}{lll}
\multicolumn{3}{l}{Optimizer: Adam }  \\
\multicolumn{3}{l}{Initial learning rate: 0.055330 } \\
\multicolumn{3}{l}{Learning rate decay factor: 0.250831 } \\ \hline
\multicolumn{1}{|l|}{Node index} & \multicolumn{1}{l|}{Input Node(s)} & \multicolumn{1}{l|}{Node type} \\ \hline
\multicolumn{1}{|l|}{0} & \multicolumn{1}{l|}{} & \multicolumn{1}{l|}{Input node} \\ \hline
\multicolumn{1}{|l|}{1} & \multicolumn{1}{l|}{0} & \multicolumn{1}{l|}{\begin{tabular}[c]{@{}l@{}}(conv): Conv2d(6, 32, kernel\_size=(3, 3), stride=(2, 2), padding=(1, 1), bias=False)\\ (activ): Sigmoid()\\ (norm): InstanceNorm2d(32, eps=1e-05, momentum=0.1, affine=False, track\_running\_stats=False)\end{tabular}} \\ \hline
\multicolumn{1}{|l|}{2} & \multicolumn{1}{l|}{1} & \multicolumn{1}{l|}{\begin{tabular}[c]{@{}l@{}}(conv): ConvTranspose2d(32, 16, kernel\_size=(3, 3), stride=(2, 2), padding=(1, 1), output\_padding=(1, 1))\\ (activ): Sigmoid()\\  (norm): BatchNorm2d(16, eps=1e-05, momentum=0.1, affine=True, track\_running\_stats=True)\end{tabular}} \\ \hline
\multicolumn{1}{|l|}{3} & \multicolumn{1}{l|}{2} & \multicolumn{1}{l|}{\begin{tabular}[c]{@{}l@{}}(conv): ConvTranspose2d(16, 16, kernel\_size=(3, 3), stride=(2, 2), padding=(1, 1), output\_padding=(1, 1))\\ (norm): BatchNorm2d(16, eps=1e-05, momentum=0.1, affine=True, track\_running\_stats=True)\end{tabular}} \\ \hline
\multicolumn{1}{|l|}{4} & \multicolumn{1}{l|}{3} & \multicolumn{1}{l|}{\begin{tabular}[c]{@{}l@{}}(conv): Conv2d(16, 4, kernel\_size=(3, 3), stride=(1, 1), padding=(1, 1))\\ (activ): ReLU()\\ (norm): BatchNorm2d(4, eps=1e-05, momentum=0.1, affine=True, track\_running\_stats=True)\end{tabular}} \\ \hline
\multicolumn{1}{|l|}{5} & \multicolumn{1}{l|}{4} & \multicolumn{1}{l|}{\begin{tabular}[c]{@{}l@{}}(conv): ConvTranspose2d(4, 3, kernel\_size=(3, 3), stride=(2, 2), padding=(1, 1), output\_padding=(1, 1), bias=False)\\ (activ): Tanh()\\ (norm): BatchNorm2d(3, eps=1e-05, momentum=0.1, affine=True, track\_running\_stats=True)\end{tabular}} \\ \hline
\end{tabular}%
}
\end{table*}

\begin{table*}[tbh]
\caption{Parameters of the best evolved architecture for the checkerboard rendering reconstruction task.}
\label{table:checkerboard}
\resizebox{\linewidth}{!}{%
\begin{tabular}{lll}
\multicolumn{3}{l}{Optimizer: RMSprop}  \\
\multicolumn{3}{l}{Initial learning rate: 0.077964} \\
\multicolumn{3}{l}{Learning rate decay factor: 0.390523} \\ \hline
\multicolumn{1}{|l|}{Node index} & \multicolumn{1}{l|}{Input Node(s)} & \multicolumn{1}{l|}{Node type} \\ \hline
\multicolumn{1}{|l|}{0} & \multicolumn{1}{l|}{} & \multicolumn{1}{l|}{Input node} \\ \hline
\multicolumn{1}{|l|}{1} & \multicolumn{1}{l|}{0} & \multicolumn{1}{l|}{\begin{tabular}[c]{@{}l@{}}(conv): ConvTranspose2d(3, 3, kernel\_size=(3, 3), stride=(2, 2), padding=(1, 1), output\_padding=(1, 1), groups=3)\\ (activ): PReLU(num\_parameters=1)\end{tabular}} \\ \hline
\multicolumn{1}{|l|}{2} & \multicolumn{1}{l|}{0, 1} & \multicolumn{1}{l|}{add - resize to second input} \\ \hline
\multicolumn{1}{|l|}{3} & \multicolumn{1}{l|}{0, 2} & \multicolumn{1}{l|}{concat - resize to second input} \\ \hline
\multicolumn{1}{|l|}{4} & \multicolumn{1}{l|}{1, 3} & \multicolumn{1}{l|}{add - resize to second input} \\ \hline
\multicolumn{1}{|l|}{5} & \multicolumn{1}{l|}{4} & \multicolumn{1}{l|}{\begin{tabular}[c]{@{}l@{}}(conv): Conv2d(6, 3, kernel\_size=(3, 3), stride=(1, 1), padding=(1, 1), bias=False)\\ (activ): ELU(alpha=1.0)\\ (norm): BatchNorm2d(3, eps=1e-05, momentum=0.1, affine=True, track\_running\_stats=True)\end{tabular}} \\ \hline
\multicolumn{1}{|l|}{6} & \multicolumn{1}{l|}{5} & \multicolumn{1}{l|}{\begin{tabular}[c]{@{}l@{}}(conv): Conv2d(3, 3, kernel\_size=(3, 3), stride=(2, 2), padding=(1, 1), bias=False)\\ (activ): PReLU(num\_parameters=1)\\ (norm): BatchNorm2d(3, eps=1e-05, momentum=0.1, affine=True, track\_running\_stats=True)\end{tabular}} \\ \hline
\end{tabular}%
}
\end{table*}

\section{Conclusions and Future Work}\label{sec:conclusion} 
This paper proposed a novel NAS method, comprised of an expressive yet compact search space, and a simple, rapidly convergent evolutionary algorithm. Adaptive average pooling was employed to alleviate topological constraints caused by mismatching dimensions of successive convolutional blocks. The performance of the proposed method was evaluated on a variety of difficult image restoration tasks, applied to the ImageNet64x64 dataset. The performance of the discovered NN architectures was compared with a high-performance human-engineered CNN architecture. The NN architectures discovered by the proposed NAS method using only 2 GPU-hours yielded lesser, but comparable performance to the human-engineered baseline architecture. The discovered architectures were significantly smaller in size than the baseline architecture, yet yielded adequate performance, and demonstrated a diversity of topological structures. Thus, the proposed NAS method was capable of discovering compact, usable solutions under significant memory and computational time constraints.

The obvious first step in future work would be to drastically increase the computational budget. If 2 hours on a single consumer GPU can reliably yield usable results, then a large GPU cluster with a larger time budget should be able to significantly surpass the results presented in this paper. The ability of the proposed NAS method to find compact architectures makes it a good option for researchers and practitioners with limited computing resources. 

There is a vast amount of potential in applying the proposed method to other image restoration problems, such as aperture synthesis in radio astronomy, as the current solutions are severely outdated hand-designed deconvolution algorithms. Improved image restoration techniques in this domain would unlock useful scientific data.

It would also be interesting to perform an in-depth study of the best architectures evolved, as the analysis of the evolved architectures may yield significant insights about NN architectures and search spaces at large.

\bibliographystyle{IEEEtran}
\bibliography{biblio}

\end{document}